\documentclass[sigconf]{acmart}

\pdfoutput=1

\makeatletter             
\def\mdseries@tt{m}            
\makeatother                   

\usepackage{listings}
\usepackage{enumitem}
\usepackage{hyperref}
\hypersetup{
    colorlinks=true,
    linkcolor=blue,
    filecolor=red,      
    urlcolor=magenta,
    breaklinks=true,
} 
\usepackage{breakurl}

\begin{document}
\sloppy

\title{XDeep: An Interpretation Tool for Deep Neural Networks}

\author{Fan Yang, Zijian Zhang$^*$, Haofan Wang$^*$, Yuening Li, Xia (Ben) Hu}\thanks{*Authors contribute during the visiting at Texas A\&M University.}

\affiliation{
\institution{Department of Computer Science and Engineering, Texas A\&M University}
}
\email{{nacoyang,liyuening,xiahu}@tamu.edu, zijianzhang0226@gmail.com, frankmiracle@outlook.com}

\renewcommand{\shortauthors}{F. Yang et al.}

\begin{abstract}
\texttt{XDeep} is an open-source Python package developed to interpret deep models for both practitioners and researchers. Overall, \texttt{XDeep} takes a trained deep neural network (DNN) as the input, and generates relevant interpretations as the output with the post-hoc manner. From the functionality perspective, \texttt{XDeep} integrates a wide range of interpretation algorithms from the state-of-the-arts, covering different types of methodologies, and is capable of providing both local explanation and global explanation for DNN when interpreting model behaviours. With the well-documented API designed in \texttt{XDeep}, end-users can easily obtain the interpretations for their deep models at hand with several lines of codes, and compare the results among different algorithms. \texttt{XDeep} is generally compatible with Python 3, and can be installed through Python Package Index (PyPI). The source codes are available at: \texttt{\url{https://github.com/datamllab/xdeep}}.
\end{abstract}


\maketitle

\section{Introduction}

In the past decade, deep neural network (DNN) has made great achievements in lots of fields, such as computer vision~\cite{krizhevsky2012imagenet} and natural language processing~\cite{vaswani2017attention}, serving as one of the most significant momentums for the booming of artificial intelligence. Despite the superior performance, the internal sophisticated structure of DNN makes its prediction extremely hard to be interpreted, which largely impedes the trust from end-users and further limits the applications to some high-stake scenarios. To alleviate such issue, increasing attentions on DNN interpretability have been paid from both industry and academia, aiming to help practitioners and researchers better understand the DNN model and its corresponding decisions. 

According to the interpretation scope, DNN interpretability can be generally achieved through two types of explanations (i.e., \emph{global} explanation and \emph{local} explanation), which shed lights on the system behaviour respectively from the model and instance level. Global explanation typically focuses on the model itself and aims to demonstrate the overall working mechanism as well as decision patterns of DNN, while local explanation specifically pays attention on particular instances and tries to explain why certain decisions are made by DNN. Existing interpretation algorithms for DNN, no matter within global scope or local scope, all need extra efforts to be fully implemented, which makes human developers inconvenient to interpret the trained DNN directly. Thus, an effective and easy-to-use interpretation tool for DNN is needed to facilitate the academic research as well as the product development. 

However, developing such an interpretation tool is non-trivial, and there are several challenges listed as follows. First, human developers usually employ different architectures of DNN for different tasks, which typically requires the interpretation tool to be architecture-agnostic regarding to various types of data. Second, it is also desired that the interpretation tool does not affect the original DNN performance, so that human developers can interpret the model or decision with no extra cost. Third, according to the diversified needs from developers, a comprehensive set of interpretation function is preferable to provide both local and global explanation under a unified API framework. 

Considering the gap between human developers and deep models, as well as those challenges mentioned, we develop an open-source Python toolbox \textbf{\texttt{XDeep}} to help users easily interpret their trained DNNs at hands, which integrates a wide range of post-hoc interpretation algorithms with the well-documented APIs. \texttt{XDeep} is capable of providing both local and global interpretation for DNN, and it sets up a clear structure for all included algorithms inside the package. As for the local interpretation part, \texttt{XDeep} covers two primary types of methodology from the state-of-the-art (i.e., the \emph{gradient-based} and \emph{perturbation-based} method). As for the global interpretation part, \texttt{XDeep} provides APIs for developers to directly interpret the DNN components, including \emph{filter}, \emph{layer} and \emph{logit}, and further visualize the global features learned by DNN. With the aid of \texttt{XDeep}, developers can easily obtain the interpretation through several lines of code, simply with a proper \emph{explainer} object. The technical contributions of \texttt{XDeep} can be summarized as follows:

\begin{itemize}[leftmargin=*]
\item A unified framework is developed for both local and global interpretation, so as to handle diversified needs from developers;
    
\item Common interpretation algorithms are implemented with well-documented APIs, which makes them easy to be employed;
    
\item Two novel algorithms (i.e., Score-CAM and CLE) are proposed and incorporated, which help generate better interpretation. 
\end{itemize}

\section{XDeep Design}

\begin{figure}
  \includegraphics[width=\linewidth]{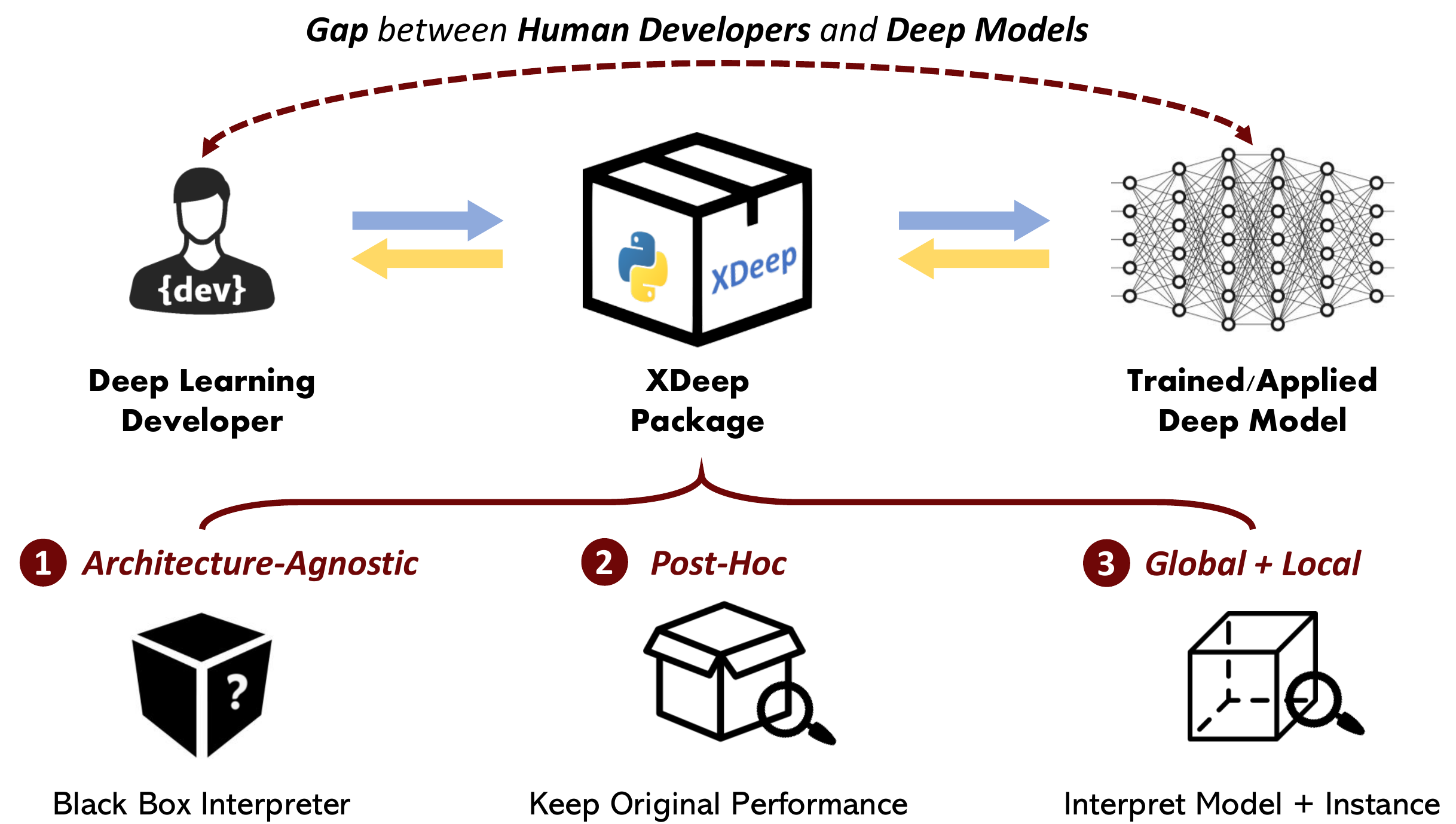}
  \caption{The overview of XDeep package.}
  \label{fig:xdeep}
\end{figure}

In this section, we introduce the overall design of \texttt{XDeep}, including the package properties, implemented algorithms and library structure. The overview of \texttt{XDeep} toolbox is shown in Figure~\ref{fig:xdeep}.

\subsection{Package Properties}

In order to effectively bridge the gap between human developers and deep models in practice, \texttt{XDeep} focuses on the following three aspects in generating interpretations for DNN.

\subsubsection{Architecture-Agnostic Interpretation}
\texttt{XDeep} aims to generate DNN interpretation in a general way, where developers can employ a same set of APIs for interpretation regardless of the DNN architecture. In practical scenarios, it is common that different developers would have different network architectures for a same task. Take the image classification for example. Some developers may use AlexNet~\cite{krizhevsky2012imagenet} under specific settings, while other developers may prefer VGGNet~\cite{Simonyan15} in another scenario. Conducting interpretation in an architecture-agnostic way would significantly simplify the communications between human developers and deep models.

\subsubsection{Post-Hoc Interpretation}
\texttt{XDeep} follows the post-hoc manner in interpreting DNN, so that developers can keep the original DNN performance and do not need extra efforts in modifying architecture for interpretability. Post-hoc interpretation mechanism also guarantees that \texttt{XDeep} can be regarded as an relatively independent module developed along with those popular deep learning frameworks (e.g. \emph{TensorFlow} and \emph{PyTorch}). In this way, developers would directly benefit with the DNN interpretability from \texttt{XDeep} package when deploying deep models for their relevant tasks. With the post-hoc interpretation, human developers can have a sense on whether DNN is properly trained, and further refine the model based on the guidance from \texttt{XDeep}. 

\subsubsection{Global + Local Interpretation}
\texttt{XDeep} is designed to provide both global and local interpretation for DNN under a unified library framework. The comprehensive interpretation APIs in \texttt{XDeep} allow developers to interpret DNN from different scopes as they need, and help form a rather complete view for users to better understand the constructed DNN. This kind of API design also strengthens the connections between the global and local interpretation, which makes it possible for practitioners to further validate the obtained interpretation from different perspectives and for researchers to thoroughly investigate the relation between them. Besides, the joint capability in both global and local interpretation ensures an extensive application scenario for \texttt{XDeep} package.

\subsection{Implemented Algorithms}
In \texttt{XDeep} package, we integrate a wide range of post-hoc interpretation algorithms for DNN interpretability from the state-of-the-art, which are introduced and organized in Table~\ref{tab_algo}. Overall, our implemented algorithms are generally divided into the \emph{Local} and \emph{Global} category based on their interpretation scope, and the \emph{Local} category is further differentiated according to the methodology particularly employed. Since gradients in DNN do not have semantic meanings for textual and tabular data, those gradient-based interpretation algorithms are typically applied to tasks related to image data. As for the perturbation-based interpretation algorithms, they are generally applied to different types of data, including image, text and table, because the relevant explanations are generated by directly interfering the input space. Although Table~\ref{tab_algo} cannot cover all existing efforts, the implemented algorithms in \texttt{XDeep} are the relatively representative ones for post-hoc DNN interpretability, where most of them have been effectively used in practice or heavily cited in research. In the later versions of \texttt{XDeep} package, we will also keep incorporating those new novel interpretation algorithms gradually. 

\begin{table}[t]
\caption {Implemented Algorithms in \texttt{XDeep}}
\vspace{-0.3cm}
\begin{tabular}{|c|c|c|c|}
\hline
\textbf{Algorithm} & \textbf{Scope} & \textbf{Method} & \textbf{Applied Data} \\ \hline
Grad-CAM~\cite{selvaraju2017grad}           & Local          & Grad             & Image                 \\
Grad-CAM++~\cite{chattopadhay2018grad}         & Local          & Grad             & Image                 \\
Score-CAM~\cite{wang2019score}          & Local          & Grad             & Image                 \\
Vanilla-BP~\cite{Simonyan2013DeepIC}         & Local          & Grad             & Image                 \\
Guided-BP~\cite{Springenberg2014StrivingFS}          & Local          & Grad             & Image                 \\
SmoothGrad~\cite{smilkov2017smoothgrad}        & Local          & Grad             & Image                 \\
IntegrateGrad~\cite{sundararajan2017axiomatic}     & Local          & Grad             & Image                 \\ \hline
LIME~\cite{ribeiro2016should}               & Local          & Pert         & Image, Text, Table      \\
Anchors~\cite{ribeiro2018anchors}            & Local          & Pert         & Image, Text, Table      \\
SHAP~\cite{lundberg2017unified}               & Local          & Pert         & Image, Text, Table      \\
CLE~\cite{zhang2019contextual}                & Local          & Pert         & Image, Text, Table      \\ \hline
Max-Activation~\cite{yosinski2015understanding}     & Global         & Grad             & Image                 \\
Invert-Feature~\cite{mahendran2015understanding}   & Global         & Grad             & Image                 \\ \hline
\end{tabular}
\label{tab_algo}
~\\
\leftline{\emph{*Grad $\longrightarrow$ Gradient-based; \quad Pert $\longrightarrow$ Perturbation-based}}
\end{table}

\subsection{Library Structure}

\begin{figure*}
  \includegraphics[width=\linewidth]{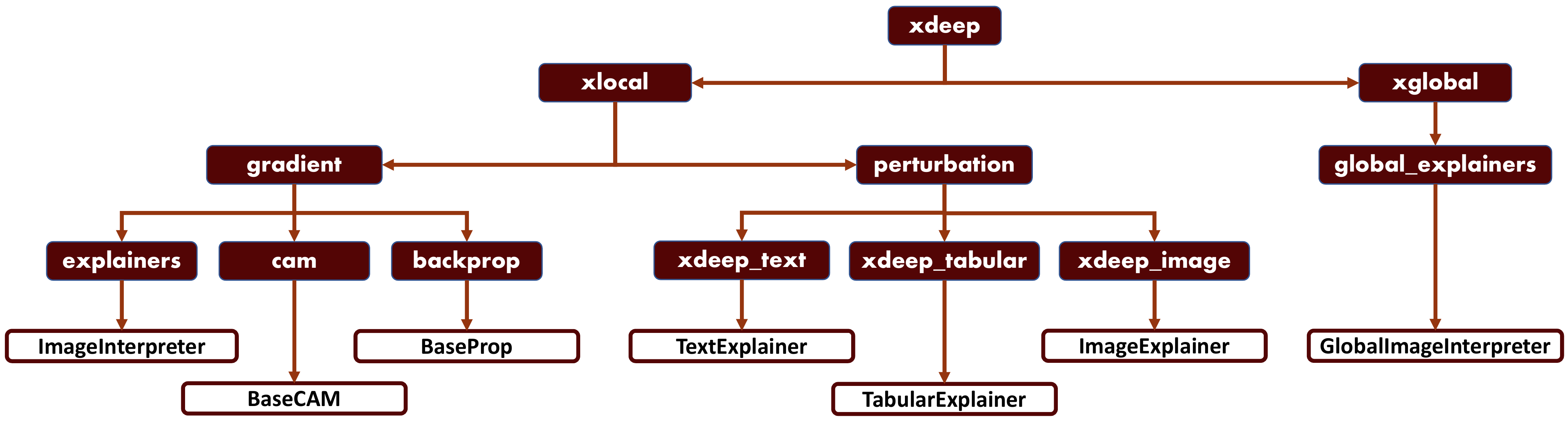}
  \caption{The library structure of XDeep package.}
  \label{fig:api}
\end{figure*}

\texttt{XDeep} is friendly to deep learning developers, since its library structure is clear and easy-to-follow. The overall structure of \texttt{XDeep} is illustrated by Figure~\ref{fig:api}, showing the high-level classes built in the package. To obtain the gradient-based local interpretation, developers simply need to construct an explainer object, instantiated from class \texttt{ImageInterpreter}, to assess all relevant algorithms and utilities. Besides, we also provide APIs for CAM and back-propagation methodologies (i.e., \texttt{BaseCAM} and \texttt{BaseProp}) individually, so as to aid developers to desgin their new algorithms for interpretation. As for the perturbation-based local interpretation, the explainer object is built regarding to different data types, which are accordingly instantiated from class \texttt{TextExplainer}, \texttt{TabularExplainer} and \texttt{ImageExplainer}. The global interpretation is further achieved by the explainer object from class \texttt{GlobalImageInterpreter}, which integrates several different methods applied on different model components. The complete documentation for \texttt{XDeep} is available at: \url{https://www.x-deep.org/}. 

\section{XDeep Demonstration}

In this section, we demonstrate how to use \texttt{XDeep} package, along with the sample code snippets. The overall user logic of \texttt{XDeep} is simple, including three steps in total: (1) load the input; (2) build the explainer; and (3) generate the interpretation. More details can be found in the tutorials on our GitHub repository. 

\subsection{Gradient-Based Local Interpretation}
\begin{figure*}
  \includegraphics[width=\linewidth]{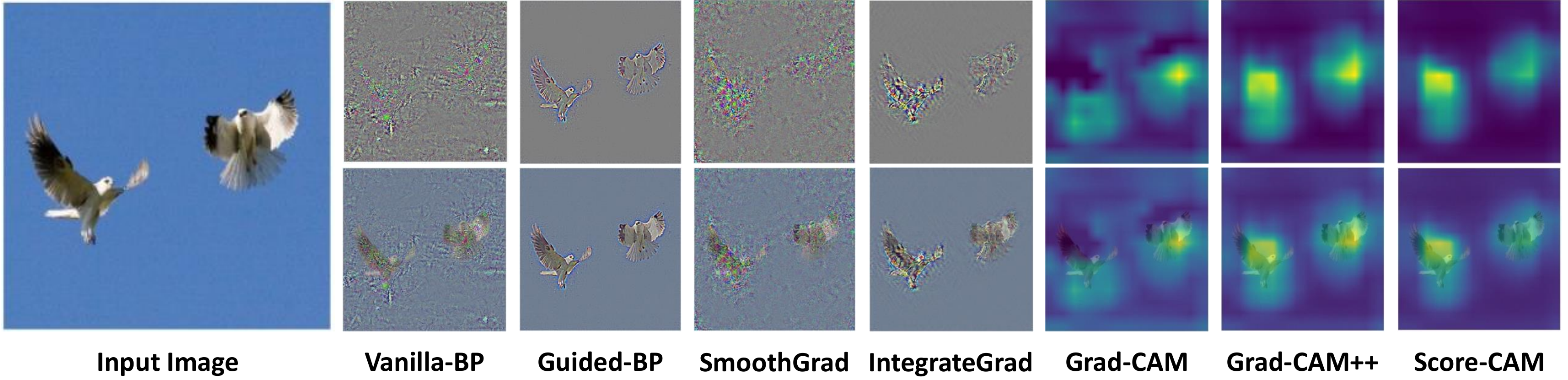}
  \caption{Visualizations of local gradient-based interpretation for VGG16 from XDeep.}
  \label{fig:local-gradient}
\end{figure*}

We first focus on the gradient-based local interpretation. Particularly, we consider the image classification with a pretrained VGG16 model, for example. The following sample codes illustrate how to use \texttt{XDeep} to obtain interpretation with Grad-CAM algorithm.
\begin{small}
\begin{lstlisting}
import torchvision.models as models
from xdeep.xlocal.gradient.explainers import *

# load input image
image = load_image('./images/input.jpg')

# load target deep model
model = models.vgg16(pretrained=True)

# build the corresponding xdeep explainer
model_explainer = ImageInterpreter(model)

# generate the local interpretation
model_explainer.explain(image, 
                        method_name='gradcam', 
                        target_layer_name='features_29',
                        viz=True) 
\end{lstlisting}
\end{small}
From the above code snippets, we note that it is much easier for developers to derive relevant interpretation with specific algorithms by \texttt{XDeep}. Simply by modifying the attribute \texttt{method\_name} of \texttt{explain} function, developers can switch the employed interpretation algorithms straightforwardly. Besides, \texttt{XDeep} also provides other necessary attributes for explainer objects (details are available in the online documentation), which makes it possible to conduct some in-depth exploration for particular interpretation algorithms. Fig.~\ref{fig:local-gradient} shows the relevant visualization results of local gradient-based interpretation, generated from \texttt{XDeep} package, where the saliency maps and overlay figures are jointly provided for end-user reference. More results can be found in our tutorials. 

\subsection{Perturbation-Based Local Interpretation}
We then discuss the perturbation-based local interpretation, regarding to different types of data, i.e., images, texts and tables. 

\subsubsection{Image Data}
\begin{figure}
  \includegraphics[width=0.95\linewidth]{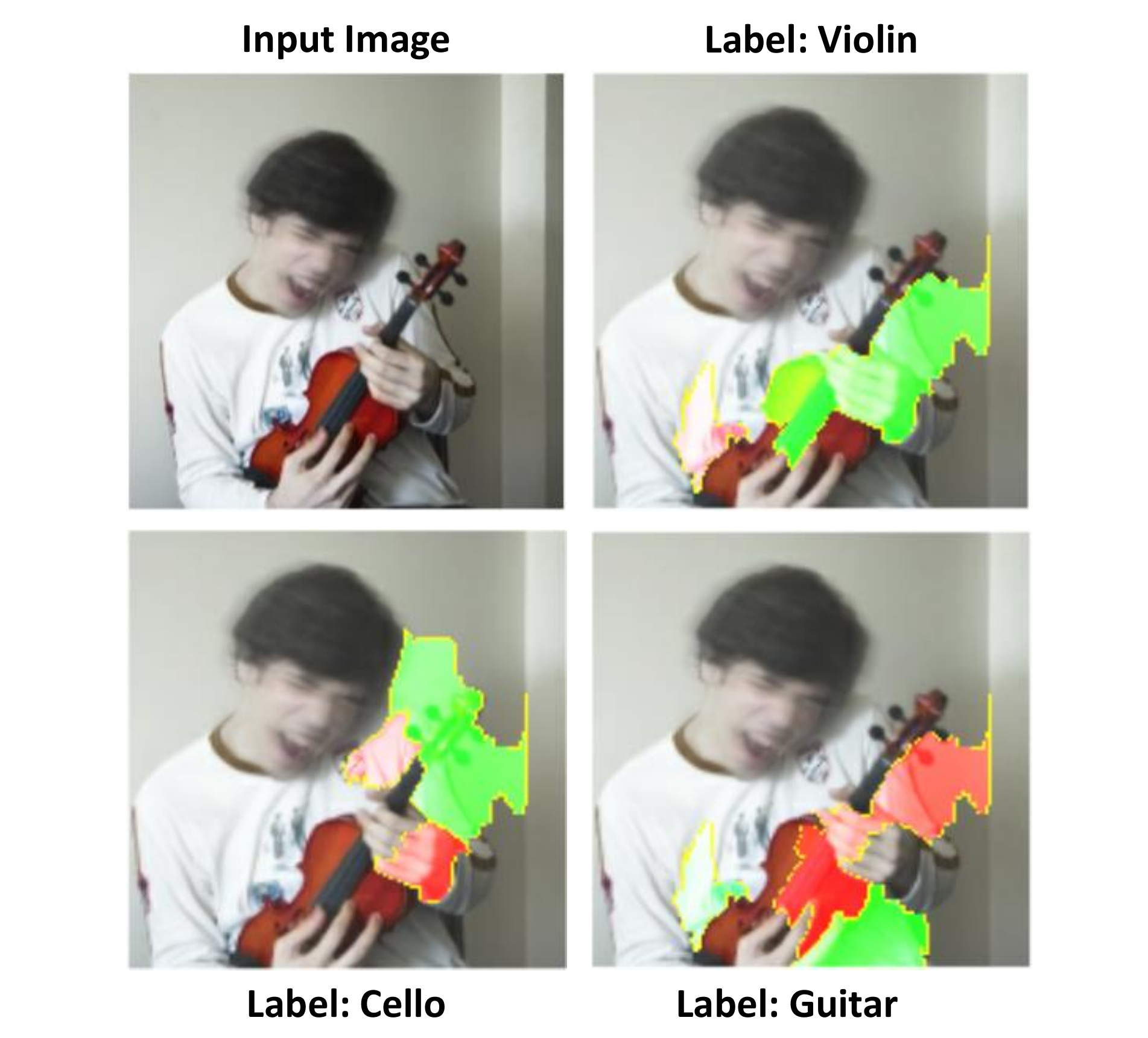}
  \caption{LIME interpretation using XDeep.}
  \label{fig:local-lime-image}
\end{figure}

As for images, perturbation-based local interpretation can be obtained through the explainer instantiated from \texttt{ImageExplainer}. The following code snippet shows how to employ the LIME algorithm with \texttt{XDeep}. The relevant visualization results are illustrated in Fig.~\ref{fig:local-lime-image}, where top 3 predictions are interpreted. 
Similarly, by using different method string (i.e., `lime'), developers can easily get the interpretations by other existing algorithms. 

\begin{small}
\begin{lstlisting}
from xdeep.xlocal.perturbation import xdeep_image

# load the input
model = load_model()
dataset = load_data()
image = load_instance()

# build the explainer
explainer = xdeep_image.ImageExplainer(model.predict, 
                                       dataset.classes)
                                       
# generate the interpretation
explainer.explain('lime', image, top_labels=3)
explainer.show_explanation('lime', deprocess=f,
                           positive_only=False)
\end{lstlisting}
\end{small}

\subsubsection{Textual Data}
\begin{figure}
  \includegraphics[width=0.95\linewidth]{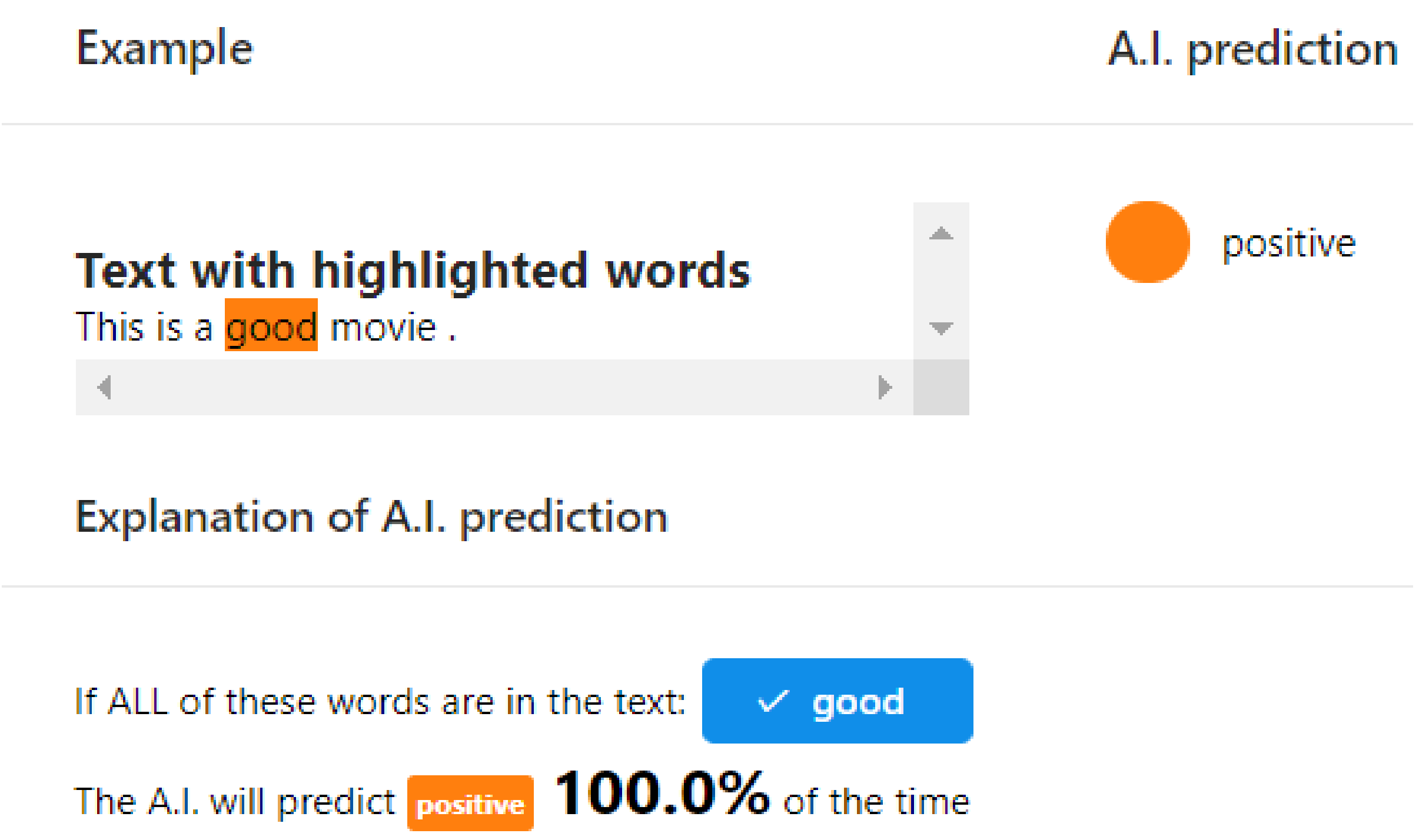}
  \caption{Anchors interpretation using XDeep.}
  \label{fig:local-anchor-text}
\end{figure}

As for texts, we build the explainer from class \texttt{TextExplainer} to obtain interpretations. The sample codes below demonstrate how to handle the textual data with \texttt{XDeep}, particularly with the algorithm Anchors. The outputs are presented by Fig.~\ref{fig:local-anchor-text}, where a simple sentiment classification is considered for illustration.

\begin{small}
\begin{lstlisting}
from xdeep.xlocal.perturbation import xdeep_text

# load the input
model = load_model()
dataset = load_data()
text = load_instance()

# build the explainer
explainer = xdeep_text.TextExplainer(model.predict,
                                     dataset.classes)
                             
# generate the interpretation
explainer.explain('anchor', text)
explainer.show_explanation('anchor')
\end{lstlisting}
\end{small}

\subsubsection{Tabular Data}
\begin{figure}
  \includegraphics[width=\linewidth]{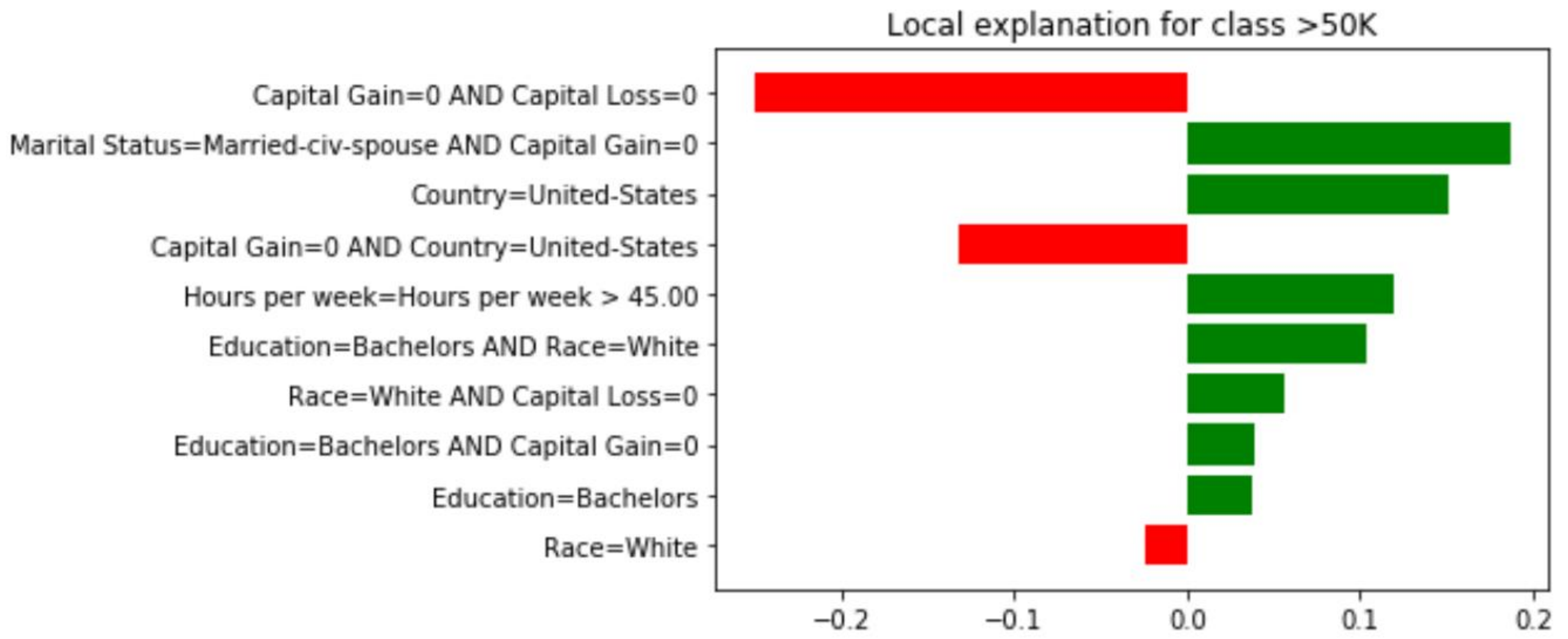}
  \caption{CLE interpretation using XDeep.}
  \label{fig:local-cle-tab}
\end{figure}

As for tables, the corresponding explainer is instantiated from class \texttt{TabularExplainer}. We consider a tabular instance classification for example, where the data comes from the \emph{Adult Data Set}\footnote{https://archive.ics.uci.edu/ml/datasets/Adult}. The following sample shows the way of \texttt{XDeep} to tackle the tabular data. The relevant results are shown in Fig.~\ref{fig:local-cle-tab}, where green bars indicate the positive feature contributions and red ones indicate the negative contributions. 
\begin{small}
\begin{lstlisting}
from xdeep.xlocal.perturbation import xdeep_tabular

# load the input
model = load_model()
dataset = load_data()
tab = dataset.test[0]

# build the explainer
explainer = xdeep_tabular.TabularExplainer(
    model.predict_proba, dataset.train,
    dataset.class_names, dataset.feature_names,
    categorical_features=dataset.categorical_features,
    categorical_names=dataset.categorical_names)

# generate the interpretation
explainer.set_parameters('cle', 
                         discretize_continuous=True)
explainer.explain('cle', tab)
explainer.show_explanation('cle')
\end{lstlisting}
\end{small}

\subsection{Global Interpretation}
\begin{figure}
  \includegraphics[width=0.8\linewidth]{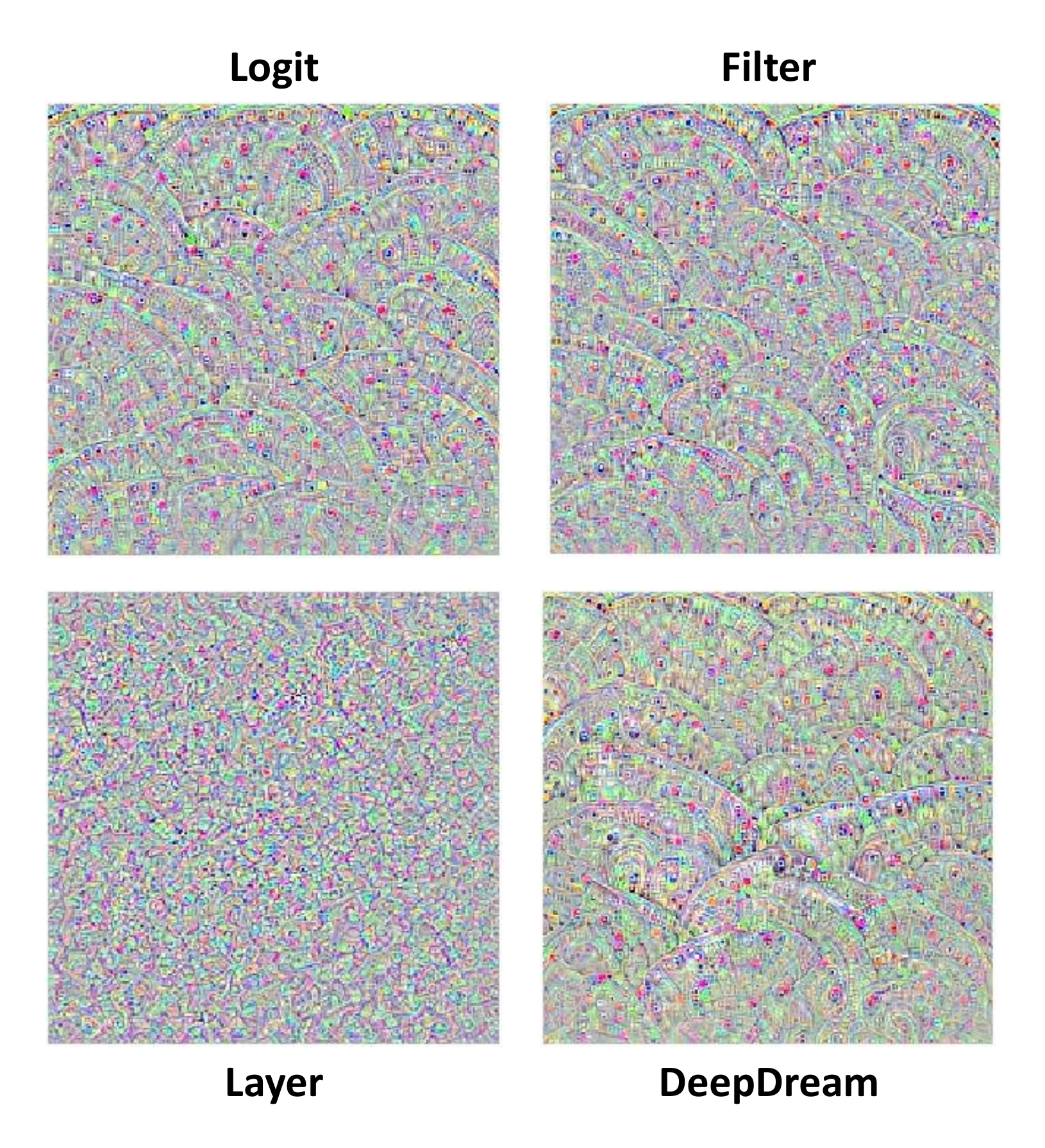}
  \caption{Visualization of VGG16 model using XDeep.}
  \label{fig:global-model}
\end{figure}

\begin{figure}
  \includegraphics[width=\linewidth]{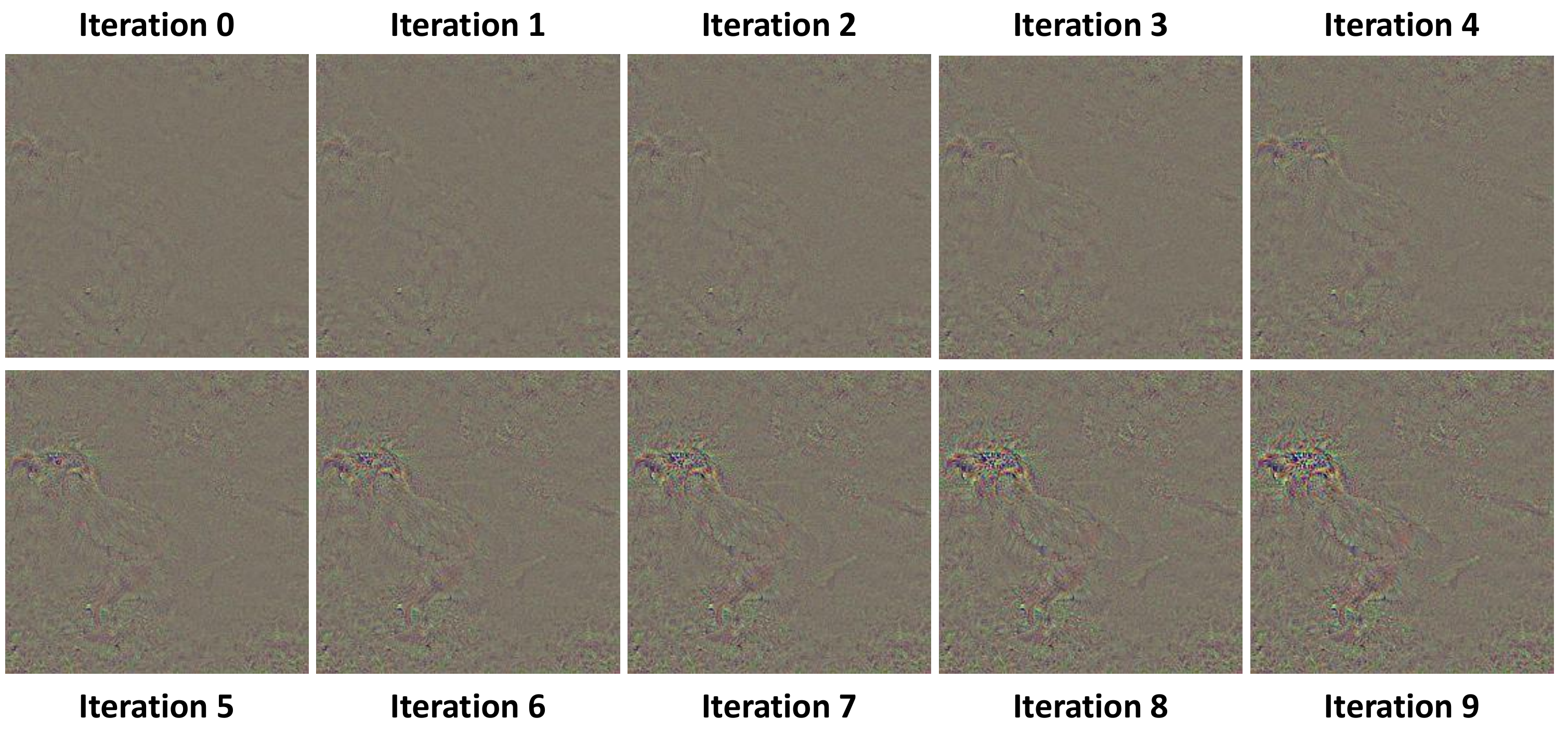}
  \caption{Visualization of inverted features using XDeep.}
  \label{fig:global-inverted}
\end{figure}

We further discuss the global interpretation, with the aid of \texttt{XDeep} package. To obtain the corresponding interpretation, explainer object from class \texttt{GlobalImageInterpreter} is needed. We demonstrate the usage of \texttt{XDeep} on global interpretation through the following sample codes. The relevant results are illustrated by Fig.~\ref{fig:global-model} and Fig.~\ref{fig:global-inverted}, where particular model components as well as global inverted features are visualized with \texttt{XDeep}. 

\begin{small}
\begin{lstlisting}
import torchvision.models as models
from xdeep.xglobal.global_explaners import *

# load the target model
model = models.vgg16(pretrained=True)

# build the explainer
model_explainer = GlobalImageInterpreter(model)

# generate global interpretation via activation maximum
model_explainer.explain(method_name='filter',
                        target_layer='features_24',
                        target_filter=20, num_iter=10)
model_explainer.explain(method_name='layer',
                        target_layer='features_24',
                        num_iter=10)
model_explainer.explain(method_name='deepdream',
                        target_layer='features_24',
                        target_filter=20,
                        input_='images/jay.jpg', 
                        num_iter=10)
                        
# generate global interpretation via feature inversion
model_explainer.explain(method_name='inverted',
                        target_layer='features_24',
                        input_='images/jay.jpg', 
                        num_iter=10)
\end{lstlisting}
\end{small}

\section{Conclusions and Future Work}

In this paper, we introduce an open-source Python package, named \texttt{XDeep}, to help deep learning developers obtain relevant interpretations from the deployed DNN. \texttt{XDeep} focuses on the architecture-agnostic post-hoc DNN interpretability, and is capable of providing both local and global interpretations. A detailed online documentation is provided for users, including all the designed classes, integrated methods, associated parameters as well as relevant tutorials. With the aid of \texttt{XDeep}, researchers and practitioners can achieve DNN interpretability simply with several lines of codes, which saves tons of efforts in employing those state-of-the-art interpretation algorithms. In the future, we will keep incorporating the new novel algorithms into \texttt{XDeep}, and try to make the package more generalizable to current mainstream deep learning frameworks, covering \emph{TensorFlow}, \emph{Keras} as well as \emph{PyTorch}. Besides, new types of interpretation will also be considered for \texttt{XDeep} improvement, such as counterfactual explanation and multi-modal explanation. 

\vspace{0.15cm}


\bibliographystyle{ACM-Reference-Format}
\bibliography{xdeep-ref}

\end{document}